# Cross-modality deep learning brings bright-field microscopy contrast to holography

Yichen Wu[1,2,3,†], Yilin Luo[1,2,3,†], Gunvant Chaudhari[4,†], Yair Rivenson[1,2,3,†], Ayfer Calis[1,2,3], Kevin De Haan[1,2,3], Aydogan Ozcan[1,2,3,4,*]

[1]Electrical and Computer Engineering Department, University of California, Los Angeles, California 90095, USA
[2]Bioengineering Department, University of California, Los Angeles, California 90095, USA
[3]California Nano Systems Institute (CNSI), University of California, Los Angeles, California 90095, USA
[4]David Geffen School of Medicine, University of California, Los Angeles, California 90095, USA
[†]Equal contribution authors     [*]Corresponding author: ozcan@ucla.edu

**Abstract:** Deep learning brings bright-field microscopy contrast to holographic images of a sample volume, bridging the volumetric imaging capability of holography with the speckle- and artifact-free image contrast of bright-field incoherent microscopy.

Digital holographic microscopy enables the reconstruction of volumetric samples from a single hologram measurement, without any mechanical scanning [1–6]. However, holographic images, for most practical applications, cannot match the speckle- and artifact-free image contrast of an incoherent bright-field microscope. Some of these holographic artifacts include twin-image and self-interference noise, which are related to the missing phase information; additional artifacts appear due to the long coherence length/diameter of the illumination source, which creates speckle and background interference from out-of-focus or unwanted objects/surfaces within the optical beam path. Stated differently, because the point spread function of a coherent imaging system has non-diminishing ripples along both the lateral and the axial directions, out-of-focus objects will create interference fringes overlapping with the in-focus objects in the holographic reconstruction, which degrades the image contrast when reconstructing volumetric samples. These issues can be *partially* mitigated using different holographic reconstruction methods, sometimes also using additional measurements [4,7–13].

Here, we use a deep neural network to perform cross-modality image transformation from a digitally back-propagated hologram corresponding to a given depth within the sample volume into an image that is equivalent to a bright-field microscope image acquired at the same depth. Because a single hologram is used to digitally propagate to different sections of the sample to virtually generate bright-field equivalent images of each section, this approach bridges the volumetric imaging capability of digital holography with speckle- and artifact-free image contrast of bright-field microscopy. After its training, the deep network learns the statistical image transformation between a holographic imaging system and an incoherent bright-field microscope. In some sense, deep learning brings together the best of both worlds by fusing the advantages of holographic and incoherent bright-field imaging modalities.

For this holographic to bright-field image transformation, we used a generative adversarial network (GAN) [14]. The network's training dataset was generated from pollen samples captured in 2D using a sticky coverslip [15] (Fig. 1). The coverslip was scanned in 3D using a bright-field microscope (Olympus IX83, 20×/0.75 NA objective lens) and a stack of 81 images with an axial spacing of 0.5 µm was captured for each region of interest, which constitute the ground truth labels. Then, in-line holograms, using a lens-less holographic microscope with a monochrome sensor [15], were acquired corresponding to the same fields-of-view (FOV) that the bright-field microscope scanned. Using a series of image registration steps, from global to local coordinates of each image patch [14], the back-propagated holograms at different depths were precisely matched to the bright-field microscope ground truth image stack in both the lateral and axial directions. These registered back-propagated holograms and the bright-field microscope image pairs were then cropped into ~10,000 patches (256 × 256 pixels) used for GAN training.

It should be emphasized that these steps need to be performed only once for the training of the GAN architecture, after which the generator network can *blindly* take as input a new hologram (never seen by the network before) to infer its bright-field equivalent image at any arbitrary depth within the sample volume. Fig. 1 demonstrates examples of these blind testing results for pollen mixtures, where the back-propagated holograms are compromised by twin-image and self-interference artifacts as well as speckle and out-of-focus interference. The generator network's output image for each case (calculated in ~0.1 sec for a FOV~0.15 mm$^2$ using a single Nvidia 1080Ti GPU) clearly shows a significantly improved contrast, free from the artifacts and noise features observed in the back-propagated holograms, matching the corresponding bright-field image (the ground truth) of the same sample depth. In addition, the deep network also correctly colorizes the output image, using an input hologram acquired with a monochrome sensor (Sony IMX219PQ,) and narrow-band illumination ($\lambda = 850\ nm$), matching the color distribution of the bright-field image as illustrated in Fig. 1 for the yellow-brown Ragweed pollens and the white Bermuda grass pollens.

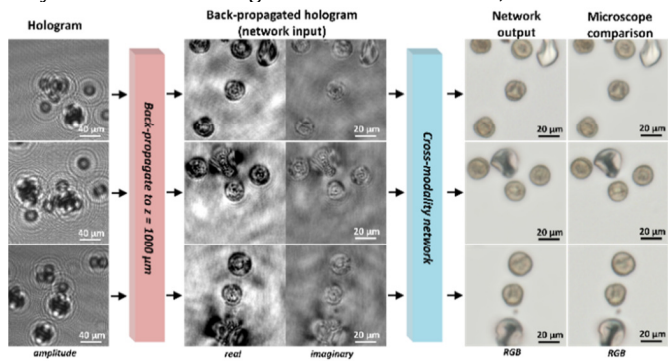

Fig. 1. Imaging of a pollen mixture captured on a sticky coverslip.

Although the deep network was only trained with pollen mixtures captured on a 2D substrate, it can successfully perform inference for volumetric imaging of samples at different depths (Fig. 2). Figs. 2(a-b) illustrate a pollen mixture captured in 3D using a polydimethylsiloxane (PDMS) substrate with ~ 800 μm thickness. A single in-line hologram of this sample (Fig. 2 (c)) was captured and numerically back-propagated (BP) to different depths within the sample volume. By feeding these back-propagated holographic images into our trained network, we once again obtained output images (Fig. 2) that are free from speckle and other interferometric artifacts observed in holography, matching the contrast and depth-of-field of bright-field microscopy images that were mechanically focused onto the same planes within the 3D sample (also see Visualization 1 for details).

This deep learning-enabled image transformation between holography and bright-field microscopy replaces the need to mechanically scan a volumetric sample, as it benefits from the digital wave-propagation framework of holography to virtually scan through the sample, where each one of these digitally propagated fields are transformed into bright-field microscopy equivalent images, exhibiting the spatial and color contrast as well as the depth-of-field expected from an incoherent microscope. In this regard, this deep learning-enabled hologram transformation network fuses the best of both worlds by bridging the volumetric digital imaging capability of holography with speckle- and artifact-free image contrast of bright-field microscopy. This can be especially useful for rapid volumetric imaging of flowing samples within a liquid [16]. In addition to in-line holography, the presented approach can also be applied to other holographic microscopy modalities to establish a statistical image transformation from one mode of coherent imaging into an incoherent microscopy modality.

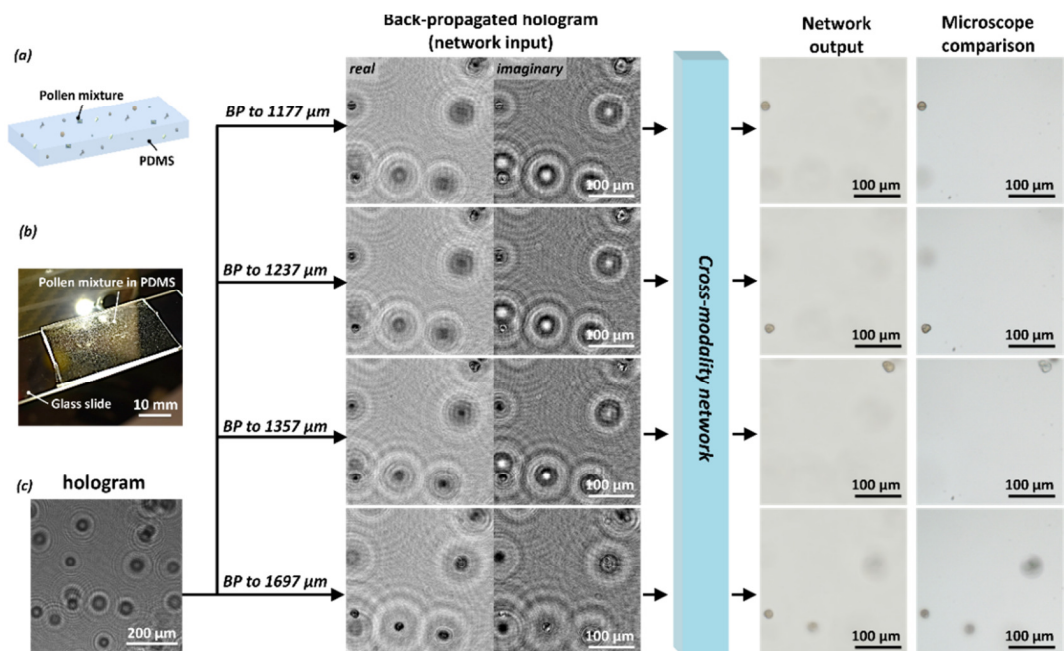

Fig. 2. Cross-modality deep learning fuses the volumetric imaging capability of holography with speckle- and artifact-free image contrast of bright-field incoherent microscopy. The pollen sample is dispersed in 3D over a PDMS substrate (thickness ~ 800 μm). BP: digital back-propagation.

**Funding.** NSF and HHMI.